\documentclass[runningheads]{llncs}
\usepackage{graphicx}
\usepackage[english]{babel}
\usepackage[T1]{fontenc}

\usepackage{amsmath}
\usepackage{graphicx}
\usepackage{amsfonts}

\usepackage[utf8]{inputenc} 
\usepackage{hyperref}       
\usepackage{url}            
\usepackage{booktabs}       
\usepackage{nicefrac}       
\usepackage{microtype}      

\usepackage{caption}
\usepackage{subcaption}
\begin{document}
\title{Solving the Real Robot Challenge using Deep Reinforcement Learning} 
%
%
\author{Robert McCarthy\inst{1}* \and
Francisco Roldan Sanchez\inst{2,3} \and
Qiang Wang\inst{1} \and David Cordova Bulens\inst{1} \and Kevin McGuinness \inst{2, 3} \and Noel O'Connor \inst{2, 3} \and Stephen Redmond \inst{1, 3}}
\authorrunning{R. McCarthy et al.}
%
\institute{University College Dublin, Ireland \and
Dublin City University, Ireland \and
Insight SFI Research Centre for Data Analytics, Ireland}
\maketitle              
\begin{abstract}

This paper details our winning submission to Phase 1 of the 2021 Real Robot Challenge\footnote{\url{https://real-robot-challenge.com} \hfill  *robert.mccarthy@ucdconnect.ie}; a challenge in which a three-fingered robot must carry a cube along specified goal trajectories. To solve Phase 1, we use a pure reinforcement learning approach which requires minimal expert knowledge of the robotic system, or of robotic grasping in general. A sparse, goal-based reward is employed in conjunction with Hindsight Experience Replay to teach the control policy to move the cube to the desired x and y coordinates of the goal. Simultaneously, a dense distance-based reward is employed to teach the policy to lift the cube to the z coordinate (the height component) of the goal. The policy is trained in simulation with domain randomisation before being transferred to the real robot for evaluation. Although performance tends to worsen after this transfer, our best policy can successfully lift the real cube along goal trajectories via an effective pinching grasp. Our approach\footnote{Code: \url{https://github.com/RobertMcCarthy97/rrc_phase1}. Videos: \url{https://www.youtube.com/playlist?list=PLLJoWXUn8XplFszi16-VZMTDBhMQFuc5o}.} outperforms all other submissions, including those leveraging more traditional robotic control techniques, and is the first pure learning-based method to solve this challenge.

\end{abstract}
%
%
%


\section{Real Robot Challenge}
\begin{figure}[h]
     \centering
     \begin{subfigure}[b]{0.48\textwidth}
         \centering
         \includegraphics[width=\textwidth]{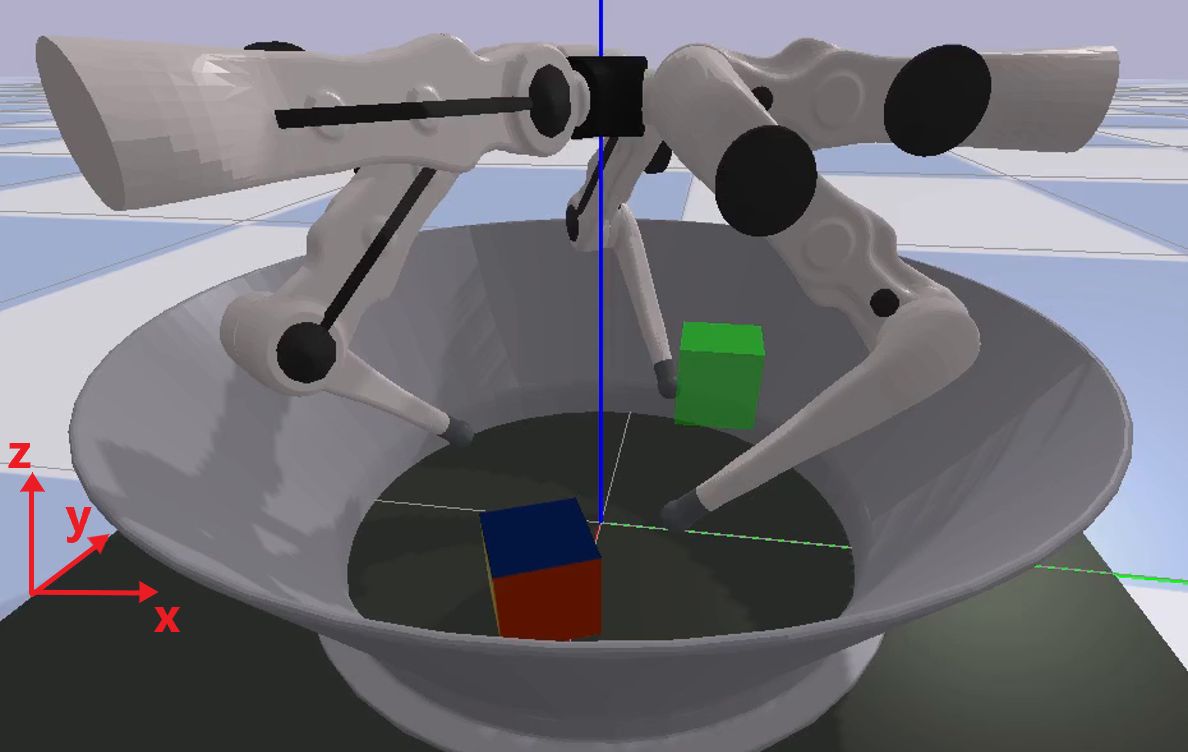}
         \subcaption{Simulation}
     \end{subfigure}
     \hfill
     \begin{subfigure}[b]{0.48\textwidth}
         \centering
         \includegraphics[width=\textwidth]{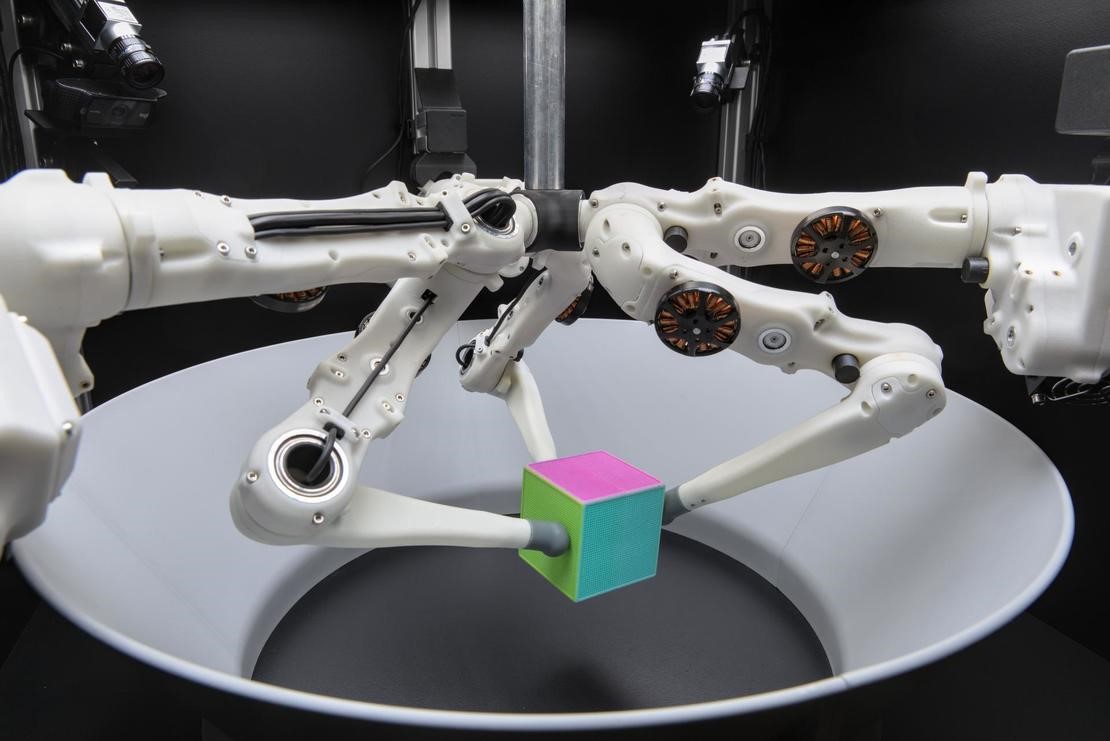}
         \subcaption{Real Robot}
     \end{subfigure}
        \caption{The simulated and real \textit{`Move Cube on Trajectory'} Trifinger robotic environments. The task is to bring the cube to specified 3-D goal coordinates, along a goal trajectory (the green cube in (a) represents a goal coordinate).}
        \label{sim-real}
\end{figure}

Dexterous robotic manipulation is applicable in various industrial and domestic settings. However, current state-of-the-art robotic control strategies generally struggle in unstructured tasks which require high degrees of dexterity. Data-driven learning methods are promising for these challenging manipulation tasks, yet related research has been limited by the costly nature of real-robot experimentation. In light of these issues, the Real Robot Challenge (RRC) \cite{rrc} aims to advance the state-of-the-art in robotic manipulation by providing participants with remote access to well-maintained robotic platforms, allowing for cheap and easy real-robot experimentation. To further support easy experimentation, users are also provided with a simulated version of the robotic setup (see Figure \ref{sim-real}).

The 2021 RRC consists of an initial qualifying Pre-Phase performed purely in simulation, followed by independent Phases 1 and 2, both performed on the real robot. Full details can be found in the `Protocol' section of the RRC website$^4$. This paper focuses solely on our approach to Phase 1 of the competition. 

In Phase 1, participants are tasked with solving the challenging \textit{`Move Cube on Trajectory'} task. In this task, a cube must be carried along a goal trajectory (which specifies the coordinates at which the cube should be positioned at each time-step) using the provided TriFinger robotic platform \cite{trifinger}. For final Phase 1 evaluation, participants submit their developed control policy and receive a score based on how well the policy can follow several randomly sampled goal trajectories.

\textit{`Move Cube on Trajectory'} requires a dexterous policy that can adapt to the various goal and cube positions encountered during an evaluation episode. Last year (2020), the winning solutions to this task consisted of structured policies that relied heavily on inductive biases and task-specific engineering \cite{dr_code,2020_winners}. We take an alternative approach, formulating the task as a pure reinforcement learning (RL) problem, and use RL to learn our control policy entirely in simulation (with domain randomisation) before transferring it to the real robot for final evaluation. Upon this evaluation, our learned policy outperformed all other competing submissions, winning Phase 1 of the 2021 RRC.

\section{Related Work}

\subsection{Traditional Robotic Manipulation}

Traditional robotic manipulation controllers often rely on solving inverse kinematic equations \cite{manipulation_review}. The goal of this approach is to find the parameters needed to position the end-effector of a robotic system (gripper, finger tips, etc.) into the desired position and orientation \cite{inverse_kin}. Because the solution to this problem is not unique, motion primitives (i.e. a set of pre-computed movements that a robot can take in a given environment) are typically introduced \cite{motion_primitives,motion_primitives2}. These primitives can each have a defined cost, allowing the robot to avoid non-smooth or non-desired transitions. Exteroceptive feedback in the form of sensors (RGB cameras, depth/tactile sensors, etc.) is usually employed to help the robot achieve the expected behaviour \cite{tactile_sensors,vision_and_force}. 

Most successful approaches in previous editions of the Real Robot Challenge make use of a combination of motion planning and motion primitives. The winning team of the 2020 edition of the challenge \cite{2020_winners} used a set of primitives to: (i) align the cube to the target position and orientation, while keeping it on the ground; and then (ii) perform grasp planning using a Rapidly-exploring Random Tree (RRT) algorithm \cite{rrt}. During the grasping planning, they use force control feedback to ensure the finger tips apply enough force to lift the cube. Finally, they improve their policy via (simulated) residual policy learning \cite{residual}, a technique that uses RL to add corrective actions to the output of the original control policy. Contrary to these methods, we use a pure learning-based approach which requires minimal task specific engineering.

\subsection{Reinforcement Learning for Robotic Manipulation}

Deep RL methods promise to allow learning of sophisticated robotic manipulation strategies that would otherwise be impossible, or at least very difficult, to hand-engineer. However, the data inefficiency of RL is a major barrier to its application in real-world robotics: real-robot data collection is time-consuming and expensive. Thus, much RL research to-date has focused on resolving or by-passing these data-efficiency issues.

Due to their generally improved sample complexity, off-policy RL methods \cite{ddpg,sac} are often preferred to on-policy methods \cite{ppo,trpo}. Model-based RL methods, which explicitly learn a model of their environment, have been proposed to further improve sample complexity \cite{pilco,mbpo,dream}, and have seen success in real robot settings (e.g., with in-hand object manipulation \cite{boadong}). Offline RL techniques seek to leverage previously collected data to accelerate learning \cite{offline}, and have learned dexterous real-world skills such as drawer opening \cite{awac}. Imitation learning methods provide the policy with expert demonstrations to learn from \cite{imitation,coarse}, enabling success in real robot tasks such as peg insertion \cite{leverage imitation}.

Finally, simulation-to-real (sim-to-real) transfer methods train a policy quickly and cheaply in simulation before deploying it on the real robot, and have famously been used to solve a Rubik's cube with a robot hand \cite{rubiks}. To account for simulator modelling errors, and to improve the policies ability to generalize to the real robot, sim-to-real approaches often employ domain randomisation \cite{sim2real,domain_randomization} or domain adaptation \cite{meta,adapt} techniques. Domain randomisation, which has been particularly effective \cite{rubiks}, randomises the physics parameters in simulation to learn a robust policy that can adapt to the partially unknown physics of the real system.

Provided with a simulated replica of the real robotic setup, but without access to prior data or expert demonstrations, we use sim-to-real transfer to bypass real-robot RL data-efficiency issues.

\section{Background}

\subsubsection{Goal-based Reinforcement Learning.}
We frame the RRC robotic environments as a Markov decision process (MDP), defined by the tuple \((\mathcal{S},\mathcal{A},\mathcal{G},p,r,\gamma,\rho_0)\). $\mathcal{S}$, $\mathcal{A}$, and $\mathcal{G}$ are the state, action and goal spaces, respectively. The state transition distribution is denoted as \(p(s'|s,a)\), the initial state distribution as \(\rho_0(s)\), and the reward function as \(r(s,g)\). \(\gamma\in(0,1)\) discounts future rewards. The goal of the RL agent is to find the optimal policy $\pi^{*}$ that maximizes the expected sum of discounted rewards in this MDP: \(\pi^{*} = \operatorname*{argmax}_\pi \mathbb{E}_{\pi} [\sum_{t=0}^{\infty} \gamma^{t} r(s_t,g_t)]\).

\subsubsection{Deep Deterministic Policy Gradients (DDPG).}

DDPG \cite{ddpg} is an off-policy RL algorithm which, in the goal-based RL setting, maintains the following neural networks: a policy (actor) \(\pi:\mathcal{S} \times \mathcal{G} \rightarrow \mathcal{A}\), and an action-value function (critic) $Q:\mathcal{S} \times \mathcal{G} \times \mathcal{A} \rightarrow \mathbb{R} $. The critic is trained to minimise the loss \(\mathcal{L}_c = \mathbb{E}(Q(s_{t},g_{t},a_{t})-y_{t})^2\), where $y_{t}=r_{t} + \gamma Q(s_{t+1},g_{t+1}, \pi(s_{t+1},g_{t+1}))$. To stabilize the critics training, the targets $y_{t}$ are produced using slowly updated polyak-averaged versions of the main networks. The actor is trained to minimise the loss: $\mathcal{L}_a = -\mathbb{E}_s Q(s,g,\pi(s,g))$, where gradients are computed by backpropagating through the combined critic and actor networks. For these updates, the transition tuples $(s_t,g_t,a_t,r_t,s_{t+1},g_{t+1})$ are sampled from a replay buffer which stores previously collected experiences (i.e., off-policy data).

\subsubsection{Hindsight Experience Replay (HER).}
HER \cite{her} can be used with any off-policy RL algorithm  in goal-based tasks, and is most effective when the reward function is sparse and binary (e.g. equation \ref{eq:xyonly}). To improve learning in the sparse reward setting, HER employs a simple trick when sampling previously collected transitions for policy updates: a proportion of sampled transitions have their goal $g$ altered to $g’$, where $g’$ is a goal achieved later in the episode. The rewards of these altered transitions are then recalculated with respect to $g’$, leaving the altered transition tuples as $(s_{t},g_{t}',a_{t},r_t',s_{t+1},g_{t+1}')$. Even if the original episode was unsuccessful, these altered transitions will teach the agent to achieve $g'$, thus accelerating its acquisition of skills.


\section{Methods}
\label{sec:method}

We train our control policy in simulation with RL before transferring it to the real robot for evaluation. This sim-to-real transfer process allows for quick and easy data collection, training, and experimentation (versus training the policy directly on the real robot). To compensate for modelling errors in the simulator, we randomise the simulation dynamics \cite{domain_randomization}. DDPG + HER is maintained as the RL algorithm, modified slightly to suit our two-component reward system. We now describe in detail our simulated environment, followed by our learning algorithm.  

\subsection{Simulated Environment}

\subsubsection{Actions and Observations.} Pure torque control of the robot arms is employed with an action frequency of 20 Hz (i.e. each time-step in the environment is 0.05 seconds). The robot has three arms and three motorised joints in each arm; thus the action space is 9-dimensional (and continuous). Observations include: (i) robot joint positions, velocities, and torques; (ii) the provided estimate of the cube's pose (i.e. its estimated position and orientation), along with the difference between the current and previous time-step's pose; and (iii) the active goal of the trajectory (i.e.  the goal coordinates at which the cube should currently be placed). In total, the observation space has 44 dimensions.


\subsubsection{Episodes.} In each simulated training episode, the robot begins in its default position and the cube is placed in a uniformly random position on the arena floor. Episodes last for 90 time-steps, with the active goal of the (randomly sampled) goal trajectory changing every 30 time-steps.

\subsubsection{Domain Randomisation.}
To help the learned policy generalize from an inaccurate simulation to the real environment, we used some basic domain randomisation (i.e., physics randomisation) during training\footnote{Our domain randomization implementation is based on the benchmark code from the 2020 RRC \cite{dr_code}.}. This includes uniformly sampling -- from a specified range -- parameters of the simulation physics (e.g. robot mass, restitution, damping, friction) and cube properties (mass and width) each episode. To account for noisy real-robot actuations and observations, uncorrelated noise is added to actions and observations within simulated episodes. See our code for more domain randomisation deails.

\subsection{Learning Algorithm}

The goal-based nature of the \textit{`Move Cube on Trajectory'} task makes HER a natural fit; HER has excelled in similar goal-based robotic tasks \cite{her} and obviates the need for complex reward engineering. As such, we use DDPG + HER as our RL algorithm\footnote{Our DDPG + HER implementation is taken from \url{https://github.com/TianhongDai/hindsight-experience-replay}, and uses hyperparameters largely based on \cite{fetch results}.}. However, in our early experiments we observed that standard DDPG + HER was slow to learn to lift the cube (especially if target goal positions were always in mid-air). To resolve this issue, we slightly alter the HER process to incorporate an additional dense reward which encourages cube-lifting behaviors, as is now described.

\subsubsection{Rewards and HER.}
In our approach, the agent receives two reward components: (i) a sparse reward based on the the cube's x-y coordinates, $r_{xy}$, and (ii) a dense reward based on the cube's z coordinate, $r_{z}$ (the coordinate frame can be seen in Figure \ref{sim-real} (a)). 

The sparse x-y reward is calculated as:

\begin{equation}
\label{eq:xyonly}
    r_{xy} =  \begin{cases} 
      0 & \textrm{if} \quad \left\| {g'}_{xy} - g_{xy} \right\|\leq 2 cm \\
     -1 & \textrm{otherwise}
   \end{cases}
\end{equation}



\noindent where ${g'}_{xy}$ are the x-y coordinates of the \textit{achieved} goal (the actual x-y coordinates of the cube), and $g_{xy}$ are the x-y coordinates of the \textit{desired} goal.

The dense z reward is defined as:
\begin{equation}
\label{eq:z}
  r_{z} =  \begin{cases} 
      - a |\, z_{cube} - z_{goal}|  & \textrm{if} \quad z_{cube} <  z_{goal} \\
      \\
      \dfrac{-a}{2}|\, z_{cube} - z_{goal}|  & \textrm{if } \quad z_{cube} > z_{goal}
   \end{cases}
\end{equation}

\noindent where $z_{cube}$ and $z_{goal}$ are the z-coordinates of the cube and goal, respectively, and $a$ is a parameter which weights $r_{z}$ relative to $r_{xy}$ (we use $a=20$).


We only apply HER to the x-y coordinates of the goal; i.e., the x-y coordinates of the goal can be altered in hindsight, but the z coordinate remains unchanged. Thus, our HER altered goals are: $\hat{g} = (g'_{x},g'_{y},g_z)$, meaning only $r_{xy}$ is recalculated after HER is applied to a sampled transition during policy updates. This reward system is motivated by the following:

\begin{enumerate}
\item Using $r_{xy}$ with HER allows the agent to learn to push the cube in the early stages of training, even if it cannot yet lift the cube to reach the z-coordinate of the goal. As the agent learns to push the cube around the x-y plane of the arena floor, it can then more easily stumble upon actions which lift it. Importantly, the $r_{xy}$ + HER approach requires no complicated reward engineering.

\item $r_{z}$ aims to explicitly teach the agent to lift the cube by encouraging minimisation of the vertical distance between the cube and the goal. It is less punishing when the cube is above the goal, serving to further encourage lifting behaviours.

\item In the early stages of training, the cube mostly remains on the floor. Consequently, during these stages most $g'$ sampled by HER will be on the floor. Thus, applying HER to $r_{z}$ here could often lead to the agent being punished for briefly lifting the cube. Since we only apply HER to the x-y coordinates of the goal, our HER altered goals $\hat{g}$ maintain their original z height. This leaves more room for the agent to be rewarded by $r_{z}$ for performing cube-lifting.
\end{enumerate}

\subsubsection{Goal Trajectories.}
In each episode, the agent is faced with multiple goals; it must move the cube from one goal to the next along a given trajectory. To ensure the HER process remains meaningful in these multi-goal episodes, we only sample future achieved goals $g'$ (to replace $g$) from the period of time in which $g$ was active.

In our implementation, the agent is unaware that it is dealing with trajectories: when updating the policy with transitions $(s_{t},g_{t},a_{t},r_{t},s_{t+1},g_{t+1})$ we always set $g_{t+1} = g_{t}$, even if in actuality $g_{t+1}$ was different\footnote{Interestingly, we found that exposing the agent (during updates) to transitions in which $g_{t+1} \neq g_{t}$ hurt performance significantly, perhaps due to the extra uncertainty this introduced to the DDPG action-value estimates.}. Thus, the policy focuses solely on achieving the current active goal and is unconcerned by any future changes in the active goal.

\subsubsection{Exploration vs Exploitation.}
We derive our DDPG + HER hyperparameters from Plappert et al. \cite{fetch results}, who use a highly `exploratory' policy when collecting data in the environment: with probability 30\% random actions are sampled (uniformly) from the action-space, and when policy actions are chosen Gaussian noise is applied. This is beneficial for exploration in the early stages of training, however, it can be limiting in the later stages when the policy must be fine-tuned -- we found that the exploratory policy repeatedly drops the cube due to the randomly sampled actions and injected action noise. To resolve this issue, rather than slowly reducing the level of exploration each epoch -- which would require a degree of hyperparameter tuning, we make efficient use of simulated evaluation episodes (performed by the standard `exploiting' policy and used to measure performance each epoch) by additionally adding them to the replay buffer. In this setup 90\% of rollouts added to the buffer are collected with the exploratory policy, and the remaining 10\% with the exploiting policy. This addition was sufficient to boost final success rates in simulation from 70-80\% to >90\%  (where "success rate" is equivalent to that seen in Figure \ref{sim train}).

\subsubsection{Full Training Process.} In practice, we first train in simulation \textit{without} domain randomisation before fine-tuning the obtained policy \textit{with} domain randomisation (DR). This was achieved as follows: (i) The policy was initially trained from scratch to convergence without DR. (ii) The replay buffer was then cleared, and the trained policy proceeded to collect several epochs of experience in the DR simulated environment. (iii) Once this initial experience was collected, normal training resumed (i.e. iterative switching between data collection and policy updates) in the DR environment. (iv) Finally, once performance converged in the DR environment, the policy was transferred to the real robot for final evaluation.

This process acted as a form of curriculum learning, allowing the policy to initially learn skills more easily in the simplified non-DR environment, before fine-tuning and `robustifying' its skills in the more challenging DR environment. We found this to improve performances both in simulation and, crucially, on the real robot. 

\section{Results}

\begin{figure}[t]
     \centering
     \begin{subfigure}[b]{0.3\textwidth}
         \centering
         \includegraphics[width=\textwidth]{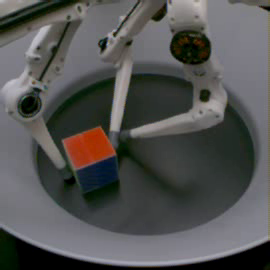}
         \subcaption{Pushing}
     \end{subfigure}
     \hfill
     \begin{subfigure}[b]{0.3\textwidth}
         \centering
         \includegraphics[width=\textwidth]{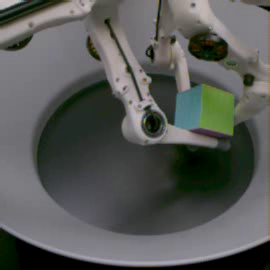}
         \subcaption{Cradling}
     \end{subfigure}
     \hfill
     \begin{subfigure}[b]{0.3\textwidth}
         \centering
         \includegraphics[width=\textwidth]{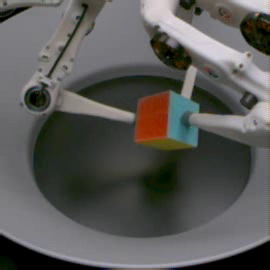}
         \subcaption{Pinching}
     \end{subfigure}
        \caption{The various manipulation strategies learned by our approach.}
        \label{fig:three graphs}
\end{figure}

\subsection{Simulation}
\begin{figure}[h]
    \centering
    \begin{subfigure}[b]{0.49\textwidth}
        \centering
        \includegraphics[width=\textwidth]{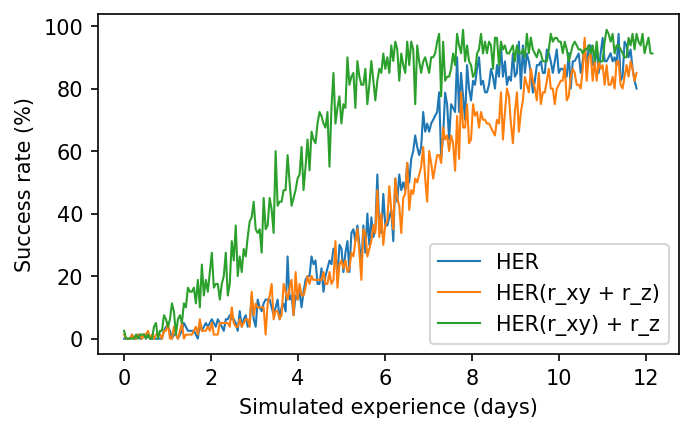}
        \caption{Simulated training}
        \label{fig:sim train}
    \end{subfigure}
    \hfill
    \begin{subfigure}[b]{0.49\textwidth}
        \centering
        \includegraphics[width=\textwidth]{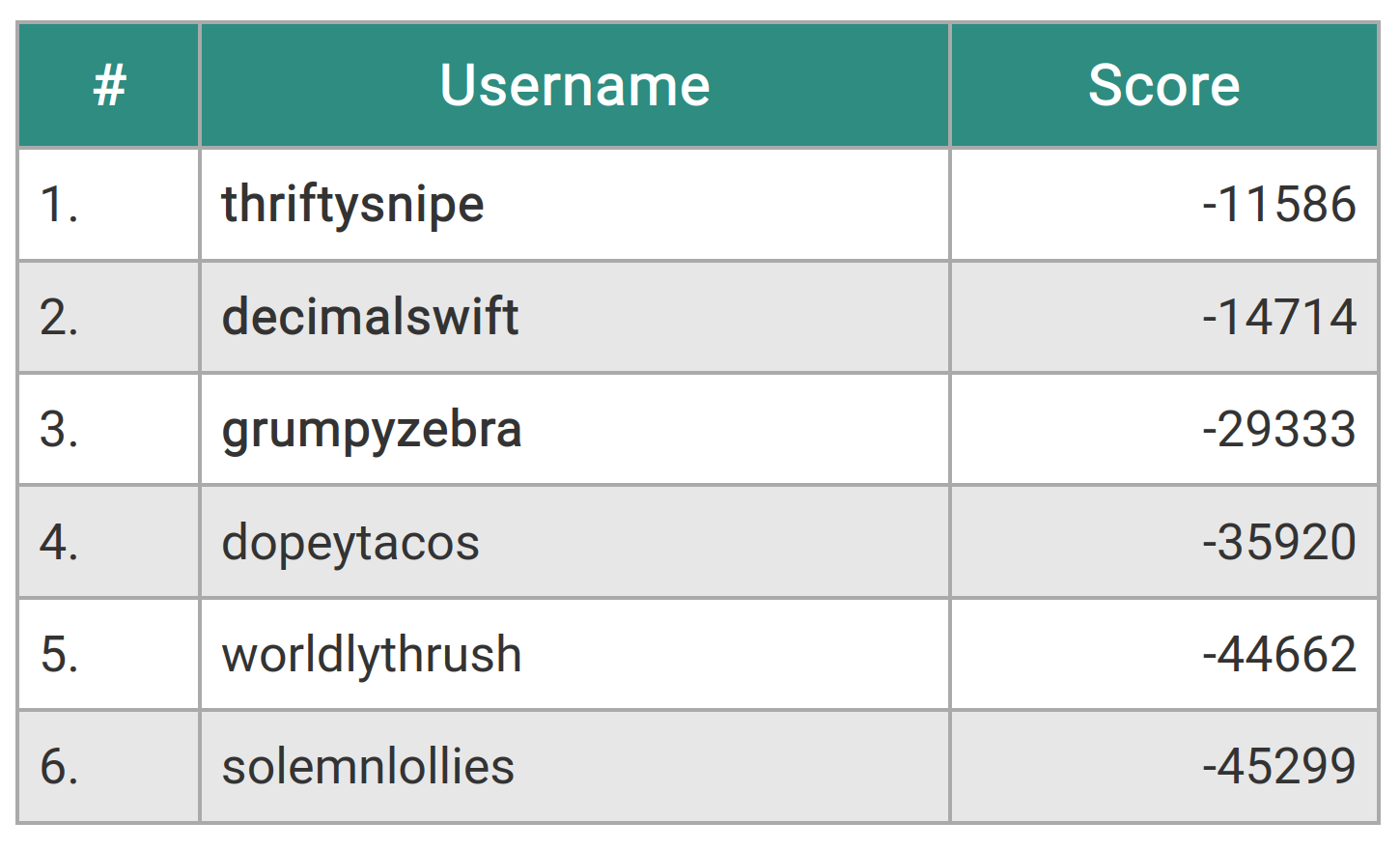}
        \caption{Final leaderboard}
        \label{fig:leaderboard}
    \end{subfigure}
    \caption{(a): Success rate vs experience collected during simulated training (1 simulated day $\approx$ 1.7 million simulated environment steps). We compare training with: (blue) HER applied to a standard sparse reward; (orange) HER applied to both $r_{xy}$ and $r_z$; and (green) our final method where HER is applied to $r_{xy}$ but not to $r_z$. An episode is deemed successful if, when complete, the final goal of the trajectory has been achieved.
    (b): The official leaderboard after the final Phase 1 RRC evaluation on the real robot (scoring system described in Table \ref{scores}). Our team name was `thriftysnipe'.
    }
    \label{sim train}
\end{figure}

Our method is highly effective in simulation. The algorithm can learn from scratch to proficiently grasp the cube and lift it along goal trajectories. Figure \ref{sim train} (a) compares the training performance of our final algorithm to that of standard HER\footnote{These runs did not use domain randomization. We trained from scratch in standard simulation before fine-tuning in a domain-randomized simulation}. Our algorithm converges in roughly $\frac{2}{3}$ the time of standard HER and is markedly improved in the the early stages of training -- this aided quicker iterative development our approach during the competition. Throughout different training runs, our policies learned several different manipulation strategies, the most distinct of which included: (i) `\textit{Pinching}' the cube with two arm tips and supporting it with the third, and (ii) `\textit{Cradling}' the cube with all three of its forearms (see Figure \ref{fig:three graphs}). 

\begin{table}[h]
  \caption{Self-reported evaluation scores of our learned \textit{Pushing}, \textit{Cradling}, and \textit{Pinching} policies when deployed on the simulated and real robots (mean $\pm$ standard deviation score over 10 episodes). Scores are based on the cumulative position error of the cube during an episode: $ score = \sum_{t=0}^{n} - (\frac{1}{2}\frac{||\textbf{e}_{xy}^{t}||}{d_{xy}} + \frac{1}{2}\frac{|e_{z}^{t}|}{d_{z}}) $, where $\textbf{e}^{t} = (e_x^t,e_y^t;e_z^t)$ is the error between the cube and goal position at time-step $t$, $d_{xy}$ is the arena range on the x-y plane, and $d_z$ the arena range on the z-axis.}
  \vspace{1.5mm}
  \label{scores}
  \centering
  \begin{tabular}{lrrr}
    \toprule
     & Pushing & Cradling & Pinching \\
    \midrule
    Simulation & -20,399$\pm$3,799 & -6,349$\pm$1,039 & -\textbf{6,198}$\pm$1,840               \\
    Real robot & -22,137 $\pm$~3,671 & -14,207 $\pm$~2,160 & \textbf{-11,489} $\pm$~3,790 \\
    \bottomrule
  \end{tabular}
\end{table}

\subsection{Real Robot}
Our final policies transferred to the real robot with reasonable success. Table \ref{scores} displays the self-reported scores of our best \textit{Pinching} and \textit{Cradling} policies under RRC Phase 1 evaluation conditions. As a baseline comparison, we trained a simple `\textit{Pushing}' policy which ignores the height component of the goal and simply learns to push the cube along the floor to the goal's x-y coordinates. The \textit{Pinching} policy performed best on the real robot, and is capable of carrying the cube along goal trajectories for extended periods of time, and of recovering the cube when it is dropped. This policy was submitted for the official RRC Phase 1 final evaluation round and obtained the winning score (see {\footnotesize \url{https://real-robot-challenge.com/2021}}, username `thriftysnipe').


The domain gap between simulation and reality was significant, and is seen to lead to inferior scores on the real robot in Table \ref{scores}. Policies often struggled to gain control of the real cube, which appeared to slide more freely than in simulation. Additionally, on the real robot policies could become stuck with an arm-tip pressing the cube into the wall. As a makeshift solution to this issue, we assumed the policy was stuck whenever the cube had not reached the goal's x-y coordinates for 50 consecutive steps, then uniformly sampled random actions for 7 steps in an attempt to `free' the policy from its stuck state.

Further relevant results, including investigations into the effects of random seeds and domain randomisation on performance, can be found in our follow-up paper \cite{qiang_paper}.

\section{Discussion}

Our relatively simple reinforcement learning approach fully solves the \textit{`Move Cube on Trajectory'} task in simulation. Moreover, our learned policies can successfully implement their sophisticated manipulation strategies on the real robot. Unlike last years benchmark solutions \cite{dr_code}, this was achieved with the use of minimal domain-specific knowledge. We outperformed all competing submissions, including those employing more classical robotic control techniques.

Due to the large domain gap, our excellent performances in simulation were not fully matched upon transfer to the real robot. Indeed, the main limitation of our approach was the absence of any training on real-robot data. It is likely that some fine-tuning of the policy on real data would greatly increase its robustness in the real environment, and developing a technique which could do so efficiently is one direction for future work. Similarly, the use of domain adaptation techniques \cite{meta,adapt} could produce a policy more capable of adapting to the real environment. However, ideally the policy could be learned from scratch on the real system; a suitable simulator may not always be available. Although our results in simulation were positive, the algorithm is somewhat sample inefficient, taking roughly 10 million environment steps to converge (equivalent to 6 days of simulated experience). Thus, another important direction for future work would be to reduce sample complexity to increase the feasibility of real robot training; perhaps achievable via a model-based reinforcement learning approach \cite{mbpo,I-HER}.

\section*{Acknowledgments}
This publication has emanated from research supported by Science Foundation Ireland (SFI) under Grant Number SFI/12/RC/2289\_P2, co-funded by the European Regional Development Fund, by Science Foundation Ireland Future Research Leaders Award (17/FRL/4832), and by China Scholarship Council (CSC). We thank the Max Planck Institute for Intelligent Systems (Stuttgart, Germany) for organizing the challenge and providing the necessary software and hardware to run our experiments remotely on a real robot. We acknowledge the Research IT HPC Service at University College Dublin for providing computational facilities and support that contributed to the research results reported in this paper.


\end{document}